\documentclass[journal]{IEEEtran}

\usepackage{amsmath,amsfonts}
\usepackage{algorithmic}
\usepackage{algorithm}
\usepackage{array}
\usepackage[caption=false]{subfig}
\usepackage{textcomp}
\usepackage{stfloats}
\usepackage{url}
\usepackage{verbatim}
\usepackage{graphicx}
\usepackage{cite}
\usepackage[bookmarksnumbered,unicode]{hyperref}
\usepackage{booktabs}
\usepackage{listings}
\usepackage{xcolor}

\begin{document}

\title{DORA: Dataflow Oriented Robotic Architecture}

\author{Xiaodong Zhang,
~Baorui Lv,~Xavier Tao,~Xiong Wang,~Jie Bao,~Yong He,~Yue Chen,~Zijiang Yang*}

\maketitle

\begin{abstract}
Robotic middleware serves as the foundational infrastructure, enabling complex robotic systems to operate in a coordinated and modular manner. In data-intensive robotic applications, especially in industrial scenarios, communication efficiency directly impact system responsiveness, stability, and overall productivity. However, existing robotic middleware exhibit several limitations: (1) they rely heavily on (de)serialization mechanisms, introducing significant overhead for large-sized data; (2) they lack efficient and flexible support for heterogeneous data sizes, particularly in intra-robot communication and Python-based execution environments. To address these challenges, we propose Dataflow-Oriented Robotic Architecture (DORA) that enables explicit data dependency specification and efficient zero-copy data transmission. We implement the proposed framework as an open-source system and evaluate it through extensive experiments in both simulation and real-world robotic environments. Experimental results demonstrate substantial reductions in latency and CPU overhead compared to state-of-the-art middleware. 
\end{abstract}

\begin{IEEEkeywords}
Software, Middleware and Programming Environments, Distributed Robot Systems, Industrial Robots, DORA
\end{IEEEkeywords}

\section{Introduction}
\IEEEPARstart{R}{obotic} middleware is introduced as a software infrastructure for unified communication interface and task assignment between different internal components or different robots\cite{ComparisonRM, RoboticFrameworks}.
Conventional robotic middlewares, such as ROS1\cite{ros1}, ROS2\cite{ros2}, and CyberRT\cite{CyberRT}, have been deployed within kinds of autonomous systems for information exchange like sensor data or actuator commands. 
However, the existing robotic middlewares often suffer the communication inefficiency due to network-based transmission mechanism, which incurs a high processing delay, especially while handling large volume of data. 

Although a variety of robotic middleware have been widely deployed in practical applications, they still exhibit several limitations in terms of communication efficiency. 
First, reliability and scalability cannot be simultaneously satisfied. 
ROS1\cite{ros1}, which depends on the stability of the master node, exhibits limited reliability\cite{zoro}. 
In contrast, ROS2\cite{ros2} and CyberRT\cite{CyberRT} are designed following a decentralized paradigm, where robotic modules rely on broadcast-based discovery to find each other; however, this approach incurs excessive communication traffic, leading to poor scalability\cite{drones9080564}\cite{zoro}. 
Second, their communication are inefficient. 
ROS1 relies on network-based transmission even for local inter-process communication, resulting in extremely high latency when handling large data payloads. 
Both ROS2 and CyberRT employ Data Distribution Service (DDS)\cite{dds1}\cite{dds2}\cite{fastdds} as their underlying communication layer, which introduces additional data-copy overhead between the robotic middleware and DDS. 
Moreover, the communication efficiency of DDS itself is often insufficient for high-performance robotic workloads\cite{oops}.

In this paper, we present Dataflow-Oriented Robotic Architecture (DORA), a general and high-efficient robotic middleware that provides a unified dataflow modeling and a significant reduction in communication overhead. 
Our key design is two-fold.

Firstly, we propose a novel robotic framework oriented to dataflow that enables efficient and secure data transmission. 
In centralized robotic middlewares, robot nodes discover each other by registering with a common server process. 
This design leads to the unreliability of robot system, as the crash of the central node can lead to the failure of the entire system\cite{zoro}.
In decentralized robotic middlewares, robot nodes discover each other through service discovery. 
However, this process significantly increases communication traffic\cite{zoro}, posing a critical scalability limitation, particularly in multi-robot and swarm robotics scenarios\cite{drones9080564}.
To solve this, DORA introduces dataflow abstraction, where the entire robot application is explicitly represented as a dataflow diagram. 

Secondly, we design an efficient shared memory communication whose core principle is to maintain consistency between the in-memory representation of communication data and its representation within shared memory. 
Our design is motivated by the observation that data's (de)serialization is only necessary for inter-robot communication, while it can be completely avoided for local communication. 
Therefore, DORA unifies the data representation such that the same structure simultaneously serves as both the storage representation and the transmission representation. 
To realize this, DORA needs to tackle two key challenges. 
First, data in robotic systems is highly diverse, requiring support for a wide variety of data formats. 
Second, in scenarios involving communication across heterogeneous robots or different programming languages, maintaining data consistency is critical to prevent parsing errors. 
To overcome these challenges, DORA employs a unified and generic data format to harmonize communication across heterogeneous contexts.
Furthermore, we develop a shared-memory allocation and reclamation algorithm that leverages the invariance of data size in robotic communication scenarios to enable partition-free and consolidation-free data transmission, thereby ensuring efficient memory utilization and low-latency communication.

We have implemented DORA based on Rust language for memory safety and performance and conducted extensive experiments across a variety of test scenarios. 
Experimental results show that, compared with the state-of-the-art, DORA reduces data transmission latency by up to 31.4$\times$, and its data (de)serialization overhead is effectively negligible. 
In real-world robotic applications, DORA can consistently maintain low transmission latency with noticeably subtle fluctuations.

\textbf{Contributions}.
\begin{itemize}
\item We introduce DORA, a high-performance robotic middleware that brings modern architectural principles into robotic systems. 
\item We propose a dataflow-oriented framework that models robotic applications as dataflow graphs. We also design a series of shared-memory–centric communication mechanisms to enable more efficient data transmission.
\item We implement DORA in Rust and its source code has been released on \href{https://github.com/dora-rs/dora}{GitHub}. We have also established a community with the goal of integrating contributions from global developers and developed a wide range of DORA-based robotic applications\cite{dora_website}. 
\item We conduct extensive experiments to demonstrate its superior efficiency and low-latency performance under various communication workloads.
\end{itemize}

\section{Background \& Motivation}
\begin{figure}
    \centering
    \includegraphics[width=\linewidth]{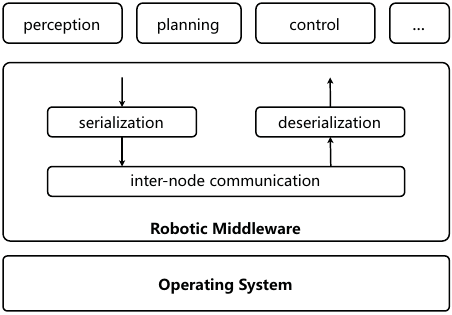}
    \caption{Inter-node communication in robotic middleware}
    \label{fig:inc}
\end{figure}

In this section, we briefly review the background of robotic middleware. 
Subsequently, we introduce two critical techniques --- (de)serialization and inter-node communication in the middleware.

\subsection{Robotic Middleware}
As robotic systems become increasingly complicated, robotic middleware acts as a critical support layer in modern robotics, bridging the gap between heterogeneous hardware, diverse software modules, and swarm robots, as shown in Figure \ref{fig:inc}. A typical robotic application consists of multiple functional units, each responsible for a specific task --- commonly represented as \textit{Node}.
E.g., a humanoid robot composed of a camera node, an algorithm node and an actuator node performs imaging, inference and action sequentially to accomplish a loco-manipulation task.
Note that these nodes are independent to ensure resource isolation and fault tolerance. 
Thus, the robotic middleware is proposed to provide efficient communication and coordination among these nodes. 

Nowadays, the predominant middleware is the ROS series which consist of ROS1\cite{ros1} (released in 2007) and ROS2\cite{ros2} (released in 2017). 
Compared to ROS1, ROS2 eliminates the long-criticized master node and instead employs a peer-to-peer discovery mechanism. 
ROS2 adopts DDS as its underlying communication layer. 
Nevertheless, poor support for Python programming language hinders the widespread application of ROS2 since it is primarily developed using C++.
Moreover, the introduction of DDS inevitably incurs multiple stages of data transmission, resulting in inefficient communication performance of ROS2. 
In addition, in order to enable mutual awareness among nodes, service discovery mechanism in ROS2 requires each node to broadcast its state to others, which significantly increases communication overhead and leads to poor scalability in complex robotic applications or multi-robot systems\cite{drones9080564}.

\textbf{Modernization and high-performance}. 
In the current era of big data–driven artificial intelligence, robotics community is in urgent need of a modern and high-performance robotic middleware.

\subsection{(De)serialization}\label{subsec:(de)serialization}
We introduce the concept of (de)serialization, which refers to the process of converting data structures to and from transmissible formats. 
As shown in Figure \ref{fig:inc}, in a typical producer–consumer communication scenario, robotic middleware serializes the data produced by upper-layer producer node into binary format and stores in shared memory; correspondingly, consumer node deserialize the associated memory regions back into the original message format.

Existing robotic middlewares widely adopt such serialization-based communication mechanisms. 
For example, ROS2 serializes messages into the Common Data Representation (CDR)\cite{fastcdr} format, while CyberRT uses Protocol Buffers (Protobuf)\cite{protobuf}. 
However, this serialization approach introduces several drawbacks. 
First, the end-to-end (de)serialization process incurs substantial computational and memory overhead, which becomes particularly pronounced for large-sized data. 
Second, for intra-robot data communication, this process is not always necessary.

\textbf{Serialization-free}. 
Complex serialization protocols can lead to long parsing times and computational overhead, which makes the (de)serialization stage a bottleneck for the entire system. 
Therefore, in robot systems, an efficient (de)serialization mechanism can significantly reduce the computational overhead caused by these operations, decrease latency, and enhance response speed.

\subsection{Inter-Node Communication}
A robot dataflow can be regarded as a directed graph, in which nodes correspond to mutually coupled robot functional nodes, and edges correspond to data communication between nodes. 
As shown in Figure \ref{fig:inc}, a core function of robotic middleware is inter-node communication.
Robotic middleware generally uses three inter-node communication mechanisms\cite{oops}: IntrA-Process communication (IAP), shared memory and socket-based method.
IAP means directly passing reference since threads share the address space of the process.
Socket-based method includes both TCP and UDP protocols, in which data to be transmitted is encapsulated into network packets and sent over the network.
Shared memory maps the same physical memory region into the virtual address spaces of multiple processes, allowing them to access the shared data directly without duplication.

Among these three approaches, shared-memory–based communication typically demonstrates the best performance\cite{oops}. 
However, FastDDS\cite{fastdds} in ROS2 supports shared-memory transport only for C++ and does not provide such support for Python. 
Moreover, in FastDDS, the allocation granularity of shared memory is fixed, meaning that data of varying sizes must be transmitted using shared-memory segments of the same size. 
This design introduces additional data partitioning overhead: if the allocated shared memory is too small, data must be fragmented, whereas if it is too large, shared memory resources are wasted.

\textbf{High-efficient allocation for shared memory}.
Robotic middleware must accommodate heterogeneous data sizes in inter-node communication to achieve efficient data transfer.

\section{DORA System Design}

\subsection{System Overview}
\begin{figure*}
    \centering
    \includegraphics[width=\linewidth]{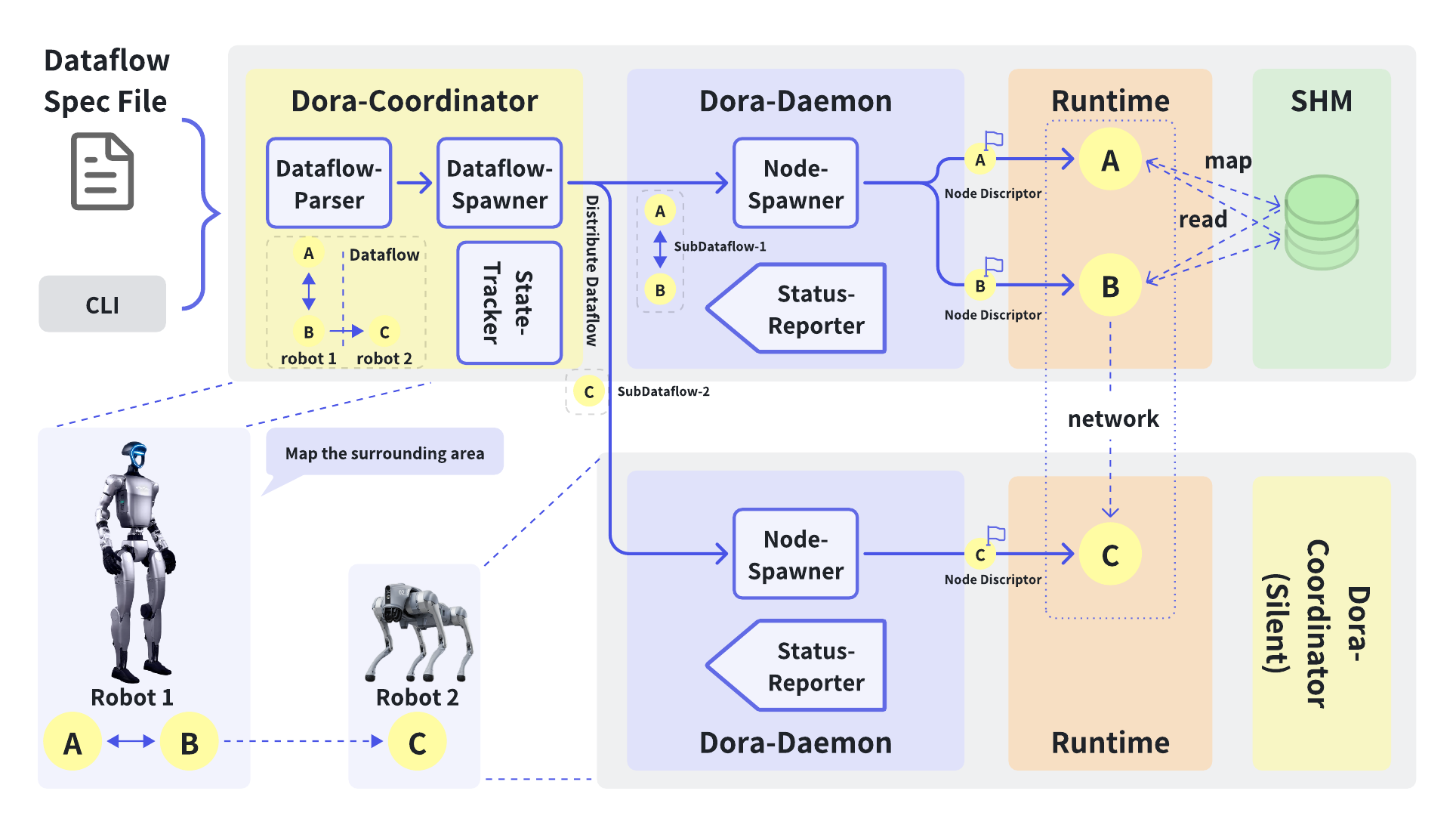}
    \caption{DORA workflow}
    \label{fig:DORA_architecture}
\end{figure*}

Figure \ref{fig:DORA_architecture} shows the architecture and the dataflow execution stpdf of DORA. 
DORA consists of two main components: the local manager \texttt{Dora-Daemon}, which is responsible for the management of all nodes within the same robot, and the global control stack \texttt{Dora-Coordinator}, which oversees dataflow deployment, inter-node communication, and lifecycle management. 
Both operate as processes that execute user commands.

For a typical dataflow consisting of nodes A, B, and C, where nodes A and B are deployed on the same robot and node C resides on another, the execution process of DORA proceeds as follows:
1) \texttt{Dora-Coordinator}, upon receiving user commands via the command line (CLI), parses the corresponding dataflow specification (will be introduced in Section \ref{subsec:dfspec}). 
Based on the node placements within the dataflow specification, it partitions the overall dataflow into multiple sub-dataflows, each to be executed on a single robot. 
These sub-dataflows are then distributed to the corresponding \texttt{Dora-Daemon} instances on each robot through Dataflow-Spawner.
2) Each \texttt{Dora-Daemon}, upon receiving its assigned sub-dataflow, instantiates the defined nodes as local processes through Node-Spawner, establishes the input/output bindings among them, and configures the necessary communication channels.
3) Once \texttt{Dora-Coordinator} receives readiness reports from all \texttt{Dora-Daemon}s, it broadcasts an \textit{AllNodesReady} signal to every \texttt{Dora-Daemon}, thereby triggering the execution of the entire dataflow.

At runtime, data produced by node A (or B) is delivered to node B (or A) via shared memory IPC, enabling serialization-free (which means that the entire data transmission process does not involve any (de)serialization operations) and low-latency transfer on the same host. 
Node B’s another output is then transmitted to node C over the network.

Throughout the execution process, each \texttt{Dora-Daemon} manages the lifecycle and resources of its local nodes and continuously reports status information to the \texttt{Dora-Coordinator}. 
In turn, \texttt{Dora-Coordinator} tracks the global state of the robotic dataflow and maintains system fault tolerance. 
This decoupled architecture --- featuring global coordination and local execution --- enables DORA to efficiently support distributed dataflows with both intra- and inter-robot communication.

\subsection{Dataflow Specification Abstraction}\label{subsec:dfspec}
We propose a DataFlow SPECification (DFspec) design to balance reliability and scalability. 
First, the system is scheduled at the granularity of dataflows, thereby eliminating single points of failure. 
Second, inter-node communication is explicitly declared within the dataflows, enabling nodes to directly discover each other without excessive state exchange. 
Finally, this design ensures isolation among dataflows across nodes, preventing mutual interference.

\begin{figure}
    \centering
    \includegraphics[width=\linewidth]{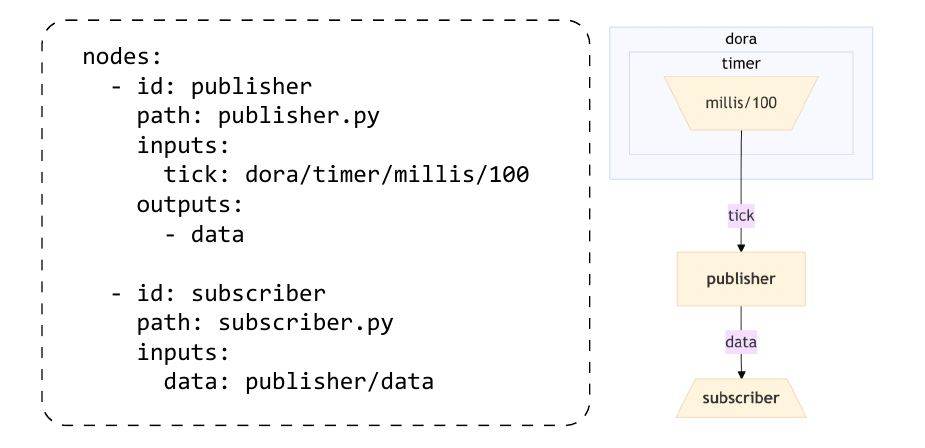}
    \caption{DORA dataflow specification}
    \label{fig:dataflow_spec}
\end{figure}

The DFspec serves as a declarative abstraction of the execution layout of robotic dataflows. 
DFspec employs a graph-based formalism, in which all entities of a dataflow are explicitly specified, including node definitions, directional data dependencies, and auxiliary system-level metadata.
A node represents the fundamental unit of data processing, whereas an edge explicitly encodes dataflow and dependency relations among nodes. 
Each node is uniquely identified and specifies input and output ports to declare its communication interfaces. 
For example, as shown in Figure \ref{fig:dataflow_spec}, the input of the subscriber node is declared as data: publisher/data, indicating that the input must originate from the data output of the publisher node within the current DFspec and is not influenced by any other dataflow or node. 
This abstraction enables DORA to decouple inter-node communication, allowing robotic application to be defined declaratively without concern for low-level mechanisms. 
Furthermore, it accommodates heterogeneous data types and transmission modes, thereby supporting various use cases such as sensor data acquisition, real-time control, and inference result dissemination.

Since the DFspec is explicitly declared prior to system execution, DORA can perform various preprocessing and optimization operations before the dataflow is launched, such as node topology sorting, communication path selection, and precomputation of resource allocation strategies. 
This design enables the system to complete part of the scheduling decisions at startup, thereby reducing runtime overhead. 
Moreover, the specification explicitly binds all relevant nodes and their communication relationships, ensuring both the integrity and consistency of data dependencies while facilitating cross-platform and cross-device deployment. 
When deploying the same dataflow to multiple robot systems, DORA only needs to distribute a unified DFspec file, from which execution plans can be quickly generated on each target system, thus enabling efficient distribution and consistency maintenance.

\subsection{Local Manager: \texttt{Dora-Daemon}}
In the DORA architecture, the core responsibility of \texttt{Dora-Daemon} is the full-lifecycle management of local sub-dataflow execution. As the local runtime control hub, \texttt{Dora-Daemon} first parses and constructs the node topology on the local host according to the DFspec and scheduling plan issued by the global coordinator (\texttt{Dora-Coordinator}), including node instantiation, binding of input/output ports, and establishment of communication channels, thereby ensuring consistency between the local execution environment and the global dataflow. 

During runtime, \texttt{Dora-Daemon} manages node start, pause, resume, and termination via event-driven mechanisms and resource scheduling. Its communication module supports high-performance local data transfer, such as zero-copy shared memory, while enabling network data exchange with remote nodes when needed. It also monitors and schedules local resources --- including CPU cores, memory, and accelerators (e.g., GPUs) --- ensuring resource isolation and performance under concurrent multi-task execution.

Through the above mechanisms, \texttt{Dora-Daemon} not only ensures efficient and stable execution of dataflows within the local scope but also provides a reliable execution foundation for global coordination in distributed systems, enabling DORA to achieve low-latency, high-reliability distributed dataflow processing in multi-robot and multi-compute node environments.

\subsection{Global Control Stack: \texttt{Dora-Coordinator}}
On each robot, \texttt{Dora-Daemon} is responsible only for the configuration, communication, and lifecycle management of local nodes. Data transfer between nodes on different robots, however, requires a higher-level manager: \texttt{Dora-Coordinator}. In the distributed architecture of DORA, \texttt{Dora-Coordinator} handles global dataflow scheduling, maintains topological consistency, and coordinates cross-host runtime management. As the global control hub, \texttt{Dora-Coordinator} first receives and parses the complete DFspec provided by the user, transforms it into a cross-host executable task graph, and partitions tasks based on node functionality, resource requirements, and communication dependencies, forming dataflow subgraphs tailored to each local host managed by \texttt{Dora-Daemon}.

During runtime, \texttt{Dora-Coordinator} maintains continuous communication with each \texttt{Dora-Daemon} through a unified control channel, periodically collecting their operational status, performance metrics, and fault reports, and conducting system health assessments from a global perspective. Additionally, \texttt{Dora-Coordinator} is responsible for maintaining dataflow topology consistency in multi-robot or multi-cluster environments, ensuring that the remaining system continues to operate according to predefined specifications even in the presence of network partitions or partial node failures.

As the global control stack, \texttt{Dora-Coordinator} not only undertakes the tasks of distributed scheduling and state synchronization but also optimizes resource utilization and system robustness from a centralized, global perspective. This design enables DORA to achieve low-latency, highly available, and dynamically adaptive distributed dataflow processing across multi-node, heterogeneous computing platforms and networked environments.

\section{Efficient Inter-node Communication}
Beyond the advantages, namely improved reliability and scalability,  brought by the dataflow abstraction, we further introduce two inter-node communication optimizations, unified memory layout and on-demand allocation and reclamation, specifically targeting shared-memory data transfer.

\subsection*{Unified Memory Layout}\label{subsec:unified}
DORA ensures that the (de)serialization overhead in inter-node communication remains minimal. 
As analyzed in Section \ref{subsec:(de)serialization}, this overhead primarily stems from the conversion between raw data and transmissible format.
For large-sized data, repeated data copying is also a non-negligible concern.
To address these issues, DORA introduces unified memory layout. 

Unified memory layout implies that data’s storage format and transmission format are identical, meaning that their in-memory representations are fully consistent. 
In other words, the binary representation of data within a robotic node is exactly the same as its representation in shared memory, and these two can be interchanged through simple memory mapping.
As a contrast, ROS2\cite{ros2} employs Common Data Representation (CDR)\cite{fastcdr} as its serialization format to ensure cross-platform compatibility. 
Essentially, ROS2 converts messages from robots with different architectures into a CDR-formatted intermediate representation. 
For the same data, ROS2 must handle data alignment and endianness, serialize the data into the CDR format, and then store it in shared memory. 
This serialization process alters the overall data structure, requiring the receiving side to retrieve the binary stream from shared memory and deserialize it to reconstruct the original data.
DORA, leveraging unified memory layout, achieves compatibility at the memory level --- it merely ensures that the in-memory representation of data is consistent across different robot architectures. 
Consequently, the sender can directly map data from its local memory into shared memory, while the receiver can access the shared memory directly as raw data without any deserialization by passing the shared memory descriptor through \texttt{Dora-Daemon}, thereby significantly reducing communication overhead.

\subsection*{On-demand Allocation and Reclamation for Shared Memory}\label{sec:shm}
DORA supports on-demand shared memory allocation, ensuring that data transmitted via shared memory is transferred completely and efficiently, without unnecessary segmentation. 
In typical robotic applications, data exchanged between nodes is often uniform in structure and size. 
For instance, in a camera node that publishes image data at a fixed frequency, the height, width, and channel count of the images remain constant, resulting in a stable data size. 
Based on this observation, DORA maintains an internal on-demand shared memory buffer queue.

When the existing shared-memory buffers cannot accommodate new data, DORA allocates a new shared memory block whose size exactly matches that of the data to be written. 
Meanwhile, DORA tracks the usage status of each shared-memory block. 
Owing to its dataflow-oriented architecture, DORA can easily determine whether a given memory block has been fully processed by all relevant receiving nodes. 
Once all reference within all receiver nodes has been destroyed, the corresponding memory block is reclaimed and returned to the buffer queue. 
During subsequent transmissions, DORA simply retrieves the smallest available block that satisfies the size requirement from the queue. 
To prevent unbounded memory growth, DORA enforces a maximum buffer capacity: when this limit is exceeded, the oldest shared-memory blocks at the head of the queue are automatically released.

In contrast, FastDDS\cite{fastdds} (chosen here as a representative example, since different versions of ROS2 use different DDS implementations) defines a fixed default shared memory size. 
When both large and small data packets are transmitted simultaneously, this fixed-size configuration can lead to inefficiencies: if the default size is too small, large data transmissions must undergo additional segmentation and reassembly operations; if the default size is too large, small data transmissions waste considerable shared-memory space. 
DORA’s on-demand allocation mechanism dynamically adjusts memory usage, achieving higher efficiency and lower resource overhead across varying data scales.

\section{Implementation}
We have implemented DORA in Rust code for the memory safety. The source code is released on \href{https://github.com/dora-rs/dora}{DORA}, which accounts for over 47, 000 lines of code.  
Its core modules, \texttt{Dora-Coordinator} and \texttt{Dora-Daemon}, together account for more than 30,000 lines. 
To enhance developer usability and enable multi-language support, we developed a command-line interface (CLI) comprising over 7,000 lines of code. 
Furthermore, to evaluate DORA’s performance, we implemented a benchmark suite written in more than 5,800 lines of Python code by referring to \cite{dora_benchmark}.

\subsection{DORA Dataflow}
We designed DORA to explicitly declare the entire dataflow in a YAML configuration file, which specifies the configurations of each node and the interfaces of data transmission between them. 
Based on this YAML file, DORA can launch the entire dataflow with a single command, including all functional nodes and their interconnections.
To further promote modularity and reuse, we established a community node hub\cite{DORA-hub} for DORA, which collects a wide variety of sensor nodes, algorithm nodes, and visualization nodes. 
These nodes are built following standardized interfaces, enabling seamless reuse across different application scenarios. 
This can greatly facilitate the rapid construction and deployment of robotic systems.
Figure \ref{fig:dataflow_spec} illustrates a simple publisher-subscriber dataflow declaration, which includes one publisher node and one subscriber node. 
An internal timer in DORA periodically triggers the publisher node to generate data, abstracted here as a single output\_id. 
The subscriber node subsequently retrieves the corresponding data based on this output\_id. 
The partial implementation of the publisher node is shown in Code \ref{fig:producer_code}.
Developers can manage and control dataflows conveniently through the command-line interface (CLI) provided by DORA. Table \ref{tab: DORA cli} summarizes the dataflow management commands and provides brief descriptions of their functions.

\begin{lstlisting}[language=Python, caption={Code of Producer Node}, captionpos=b, label={fig:producer_code}]
from dora import Node

node = Node()
for event in node:
  if event["type"] == "INPUT" and event["id"] == "tick":
    node.send_output("data", data, metadata)
\end{lstlisting}

\begin{table*}[htbp]
\centering
\caption{DORA Command Line}
\label{tab: DORA cli}
\begin{tabular}{lll}
\toprule
\textbf{Control Object} & \textbf{Command Line} & \textbf{Description} \\
\midrule
\textbf{Dataflow} &
\begin{tabular}[c]{@{}l@{}}
\texttt{new} \\
\texttt{build} \\
\texttt{run} \\
\texttt{start} \\
\texttt{stop} \\
\texttt{list} \\
\texttt{logs} \\
\end{tabular} &
\begin{tabular}[c]{@{}l@{}}
Generate a new project or node. Choose the language between Rust, Python, C or C++ \\
Run build commands provided in the given dataflow \\
Run a dataflow locally \\
Start the given dataflow path. Attach a name to the running dataflow by using --name \\
Stop the given dataflow UUID \\
List running dataflows \\
Show logs of a given dataflow and node
\end{tabular} \\
\midrule
\textbf{DORA} &
\begin{tabular}[c]{@{}l@{}}
\texttt{check} \\
\texttt{up} \\
\texttt{destroy} \\
\texttt{coordinator} \\
\texttt{daemon} \\
\texttt{graph} \\
\end{tabular} &
\begin{tabular}[c]{@{}l@{}}
Check if the coordinator and the daemon is running \\
Spawn coordinator and daemon in local mode (with default config) \\
Destroy running coordinator and daemon \\
Run coordinator \\
Run daemon \\
Generate a visualization of the given graph using mermaid.js. Use --open to open browser
\end{tabular} \\
\bottomrule
\end{tabular}
\end{table*}

\subsection{DORA Components}
For DORA’s core components, namely \texttt{Dora-Coordinator} and \texttt{Dora-Daemon}, we predefine the message structures exchanged between them and leverage the open-source projects Tokio\cite{tokio} and Futures\cite{futures} to implement TCP-based communication for control signals and small-sized data, as well as asynchronous internal computation. 
Given that \texttt{Dora-Coordinator} and \texttt{Dora-Daemon} operate in distinct control domains, DORA supports launching a single coordinator alongside multiple local daemons on different robots. 
This design enables distributed dataflow management and communication across multiple robots. 
Table \ref{tab: DORA cli} summarizes the management commands of DORA along with their brief descriptions.
For the core shared-memory communication technology, we developed a dedicated shared\_memory\_extended\cite{shared_memory} crate to enable inter-process shared-memory communication. 
Building upon this foundation, DORA achieves highly efficient shared-memory data transfer by adopting the Apache Arrow\cite{arrow} format as a unified in-memory communication schema, where data is stored in a columnar layout and transmitted across processes in the same representation.
Furthermore, we integrate Zenoh\cite{zenoh} to facilitate communication between robots and across edge–cloud infrastructures, thereby extending DORA’s applicability to a broader range of robotic collaboration and deployment scenarios.

\section{Evaluation}

\subsection{Experiment Setup}
All benchmarks are conducted on the platform equipped with AMD Ryzen5 5600 CPU, 32 GB RAM, and 4060 Nvidia GPU. 
The operating system is Ubuntu 22.04 with Linux kernel 6.8.0-85-generic. 
All Benchmarks are written by Python and the measurement is conducted with high resolution clock from time package. 

We compare DORA with three baselines, i.e., ROS1\cite{ros1}, ROS2\cite{ros2}, and CyberRT\cite{CyberRT}.
In ROS1 (Noetic), communication --- both intra-robot and inter-robot --- is implemented using socket-based mechanism. 
ROS2 (Humble) employs FastDDS\cite{fastdds} as its communication middleware, which supports shared-memory communication for C++ nodes in local environments. 
However, for Python nodes, communication still relies on socket-based transmission. 
In contrast, CyberRT (version 10.0.0) provides a Python interface that internally invokes C++ shared-memory communication primitives.

\subsection{(De)serialization Overhead}
To evaluate DORA’s performance in (de)serialization, we constructed a producer–consumer dataflow in which the producer node publishes data with size of 4 MB at a frequency of 50 Hz, while the consumer node receives that. 
We measured the CPU utilization of both nodes during communication: the producer’s CPU usage reflects the overhead of serialization, and the consumer’s CPU usage represents that of deserialization.
Since both ROS1 and ROS2 adopt socket-based mechanisms for inter-process communication in Python, we conduct experimental evaluations exclusively on ROS2.

\begin{figure}
    \centering
    \includegraphics[width=\linewidth]{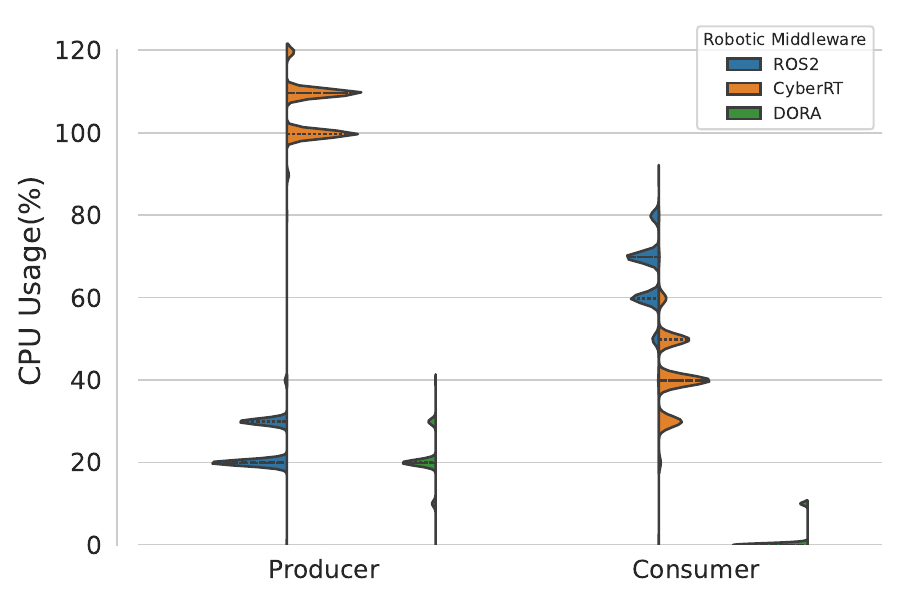}
    \caption{CPU utilization of serialization and deserialization in ROS2, CyberRT and DORA}
    \label{fig:cpu_usage}
\end{figure}

As shown in Figure \ref{fig:cpu_usage}, on the producer side, DORA’s serialization overhead is comparable to that of ROS2, since DORA must also convert raw data into the unified memory layout introduced in Section \ref{subsec:unified} before transmission.
Nevertheless, DORA’s overhead remains significantly lower than that of CyberRT, which relies on protobuf \cite{protobuf} for data formatting. 
On the consumer side, DORA allows direct access to the required data from shared memory without performing deserialization, resulting in near-zero CPU overhead. 
In contrast, both ROS2 and CyberRT must reconstruct the original data from binary sequences, leading to considerably higher CPU utilization.
CyberRT’s CPU utilization on the producer side exceeds 100\% in the figure, as the reported value represents the aggregate CPU usage of the producer process across multiple CPU cores.

\subsection{Transmission Efficiency}
To evaluate this, we conducted two experiments. 
In the first experiment, we examined the impact of different data sizes on transmission latency. 
In the second experiment, we investigated how different data publishing frequencies affect latency. 
Both local transmission (inter-process) and remote transmission (within a local area network) are evaluated. 
The experiments are conducted in a gigabit local area network (LAN) environment.

\begin{figure*}[!t]
    \centering
    \subfloat[Mean latency for small-sized data transmission\label{fig:avg_latency_small}]{
        \includegraphics[width=0.48\linewidth]{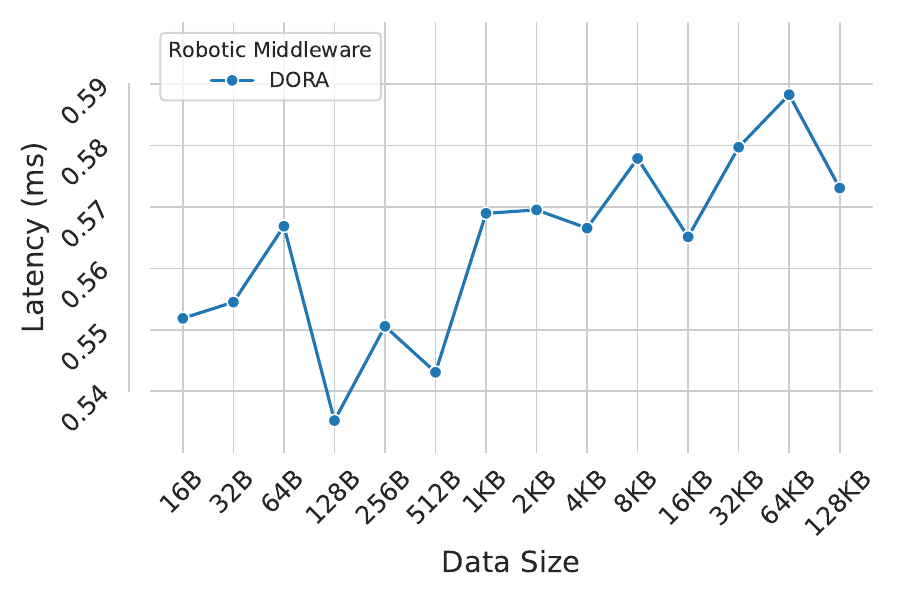}
    }
    \hfill
    \subfloat[Mean latency for large-sized data transmission\label{fig:avg_latency}]{
        \includegraphics[width=0.48\linewidth]{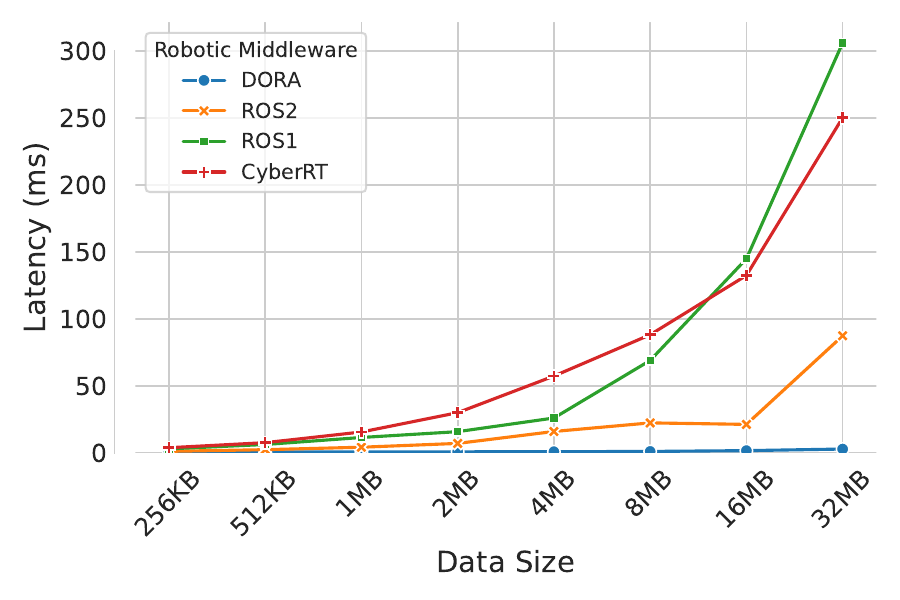}
    }
    \caption{Comparison of mean transmission latency for different data sizes}
    \label{fig:latency_compare}
\end{figure*}

The results of the first experiment are shown in Figure \ref{fig:latency_compare}. 
In this setup, a single producer and a single consumer are used, with the data publishing frequency fixed at 50 Hz, which is typical for many sensor applications. 
In Figure \ref{fig:avg_latency_small}, it can be observed that for small-sized data (e.g., a grayscale image with resolution of 360 $\times$ 360 is approximately 126 KB), DORA maintains a transmission latency within 0.59 ms, which is practically negligible. 
As the data size increases, in Figure \ref{fig:avg_latency}, DORA continues to sustain low and stable latency. 
For instance, even when transmitting 32 MB of data at 50 Hz, the latency remains around 2.78 ms.
In contrast, when the data size increases from 256 KB to 32 MB (a 128$\times$ increase), the latency of ROS1 and ROS2 rises by approximately 105$\times$ and 82$\times$, respectively. 
When transmitting 32 MB of data at 50 Hz, the latencies of ROS1 and ROS2 reach approximately 306 ms and 87 ms, respectively, which are extremely high for real-time robotic applications (a duration comparable to the inference time of several vision-language-action models).
Notably, CyberRT shows an even higher latency of 250 ms. This is because, for Python implementations, CyberRT internally invokes C++ interfaces, which introduce data copying overhead --- a process that becomes extremely time-consuming for large data transmissions.
Furthermore, we also measured the latency of CyberRT implemented in C++ for transmitting 32 MB of data, which is approximately 20 ms --- still about 7$\times$ higher than that of DORA.

\begin{figure}
    \centering
    \includegraphics[width=\linewidth]{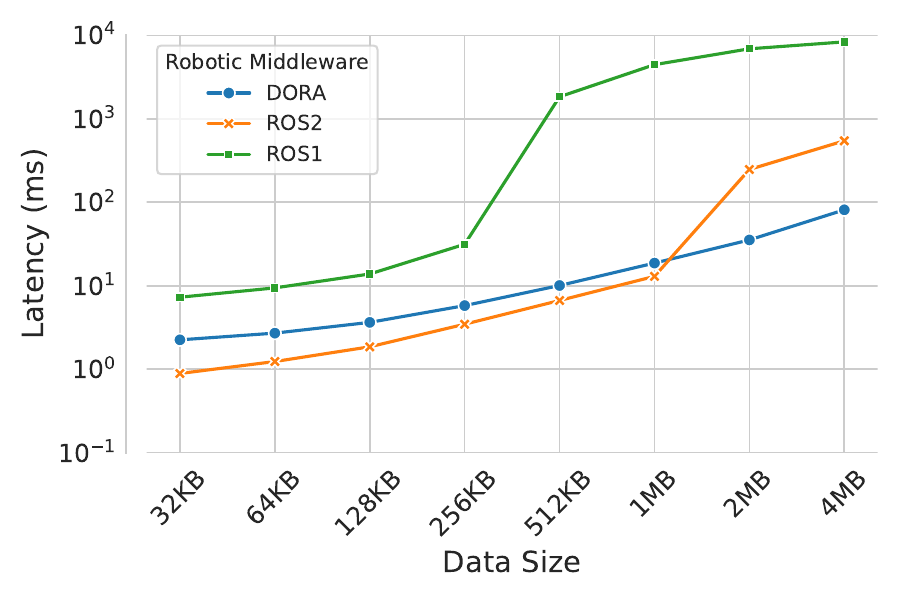}
    \caption{Comparison of mean transmission latency for different data sizes under LAN}
    \label{fig:avg_latency_remote}
\end{figure}
Figure \ref{fig:avg_latency_remote} presents a performance comparison of DORA, ROS1, and ROS2 in data transmission over a Local Area Network (LAN). 
It can be observed that all three systems exhibit a significant increase in latency starting from a data size of 32 KB. 
Specifically, as the data size increases from 32 KB to 4 MB, the latency of ROS1 rises dramatically from 7.285 ms to 8.309 s --- an increase of approximately 1, 140 times. 
For ROS2, the latency grows from 0.89 ms to 543.667 ms, corresponding to a growth factor of about 610. 
In contrast, the latency of DORA nodes increases only from 2.249 ms to 81.033 ms, representing merely a 36-fold increase.
Another noteworthy observation is that ROS1 consistently exhibits much higher overall latency compared with both DORA and ROS2. 
When the data size is below 1 MB, ROS2 nodes achieve lower transmission latency than DORA, likely benefiting from the efficiency of its underlying DDS mechanism. 
However, when the data size exceeds 1 MB, DORA demonstrates superior performance, maintaining lower latency than ROS2 and achieving the highest overall efficiency among the three systems.

\textit{Overall, when transmitting small-size data ($<$ 128 KB), DORA maintains a transmission latency of approximately 0.5 ms. 
For large-size data ($>$ 128 KB), DORA’s latency remains significantly lower than that of ROS1, ROS2, and CyberRT, achieving a 1.8 $\sim$ 31.4$\times$ reduction compared to the best-performing baseline, ROS2. 
When the scenario shifts from local transmission to LAN transmission, both ROS1 and ROS2 experience a sharp increase in latency once the data size exceeds 16 KB. 
In contrast, although DORA’s latency also increases, the growth trend remains very moderate, demonstrating its superior scalability and efficiency in networked environments.}

\begin{figure*}[!t]
\centering
\subfloat[Comparison of mean transmission latency under different data transmission frequencies\label{fig:latency_frequency_local}]{
    \includegraphics[width=0.48\linewidth]{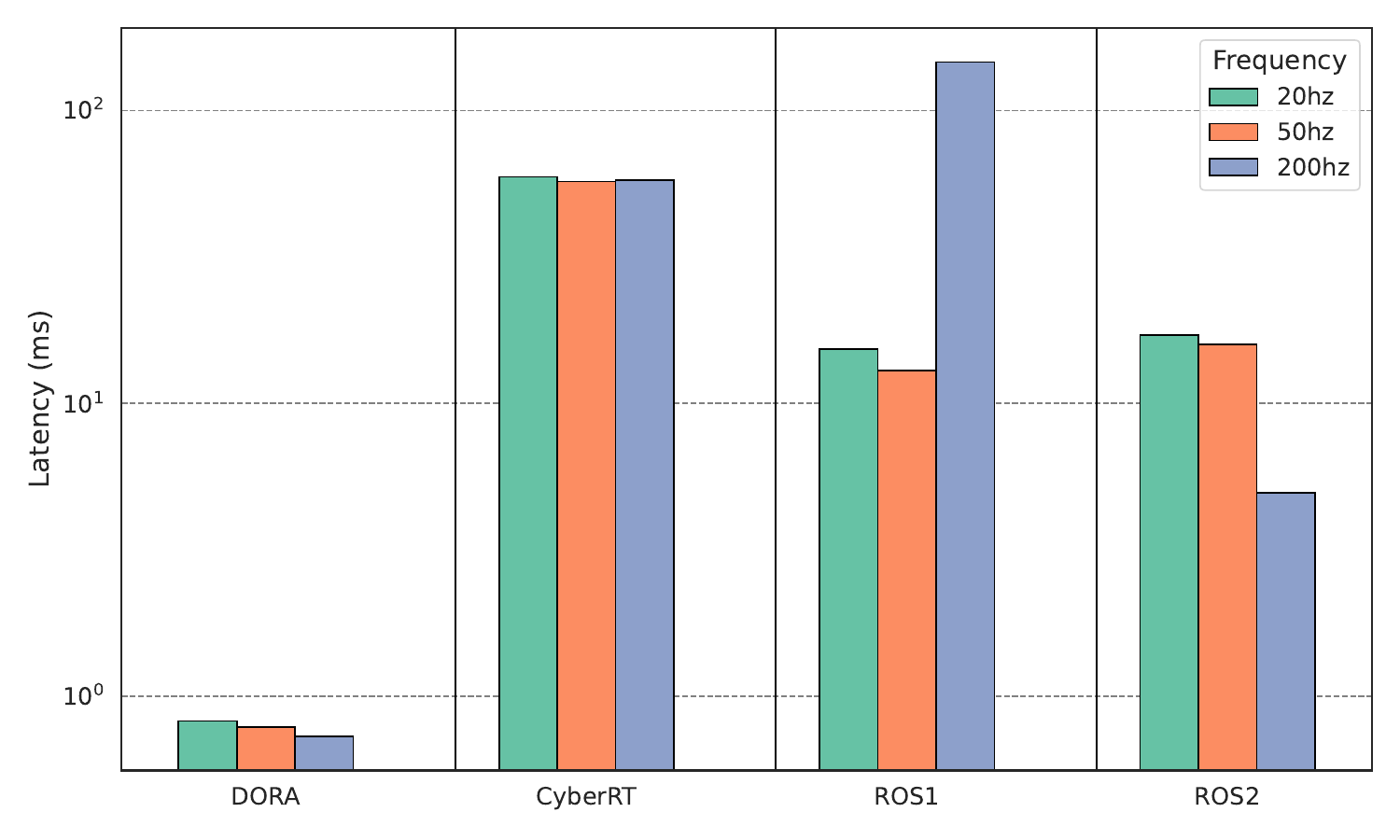}
}
\hfill
\subfloat[Comparison of mean transmission latency under different data transmission frequencies in LAN\label{fig:latency_frequency_remote}]{
    \includegraphics[width=0.48\linewidth]{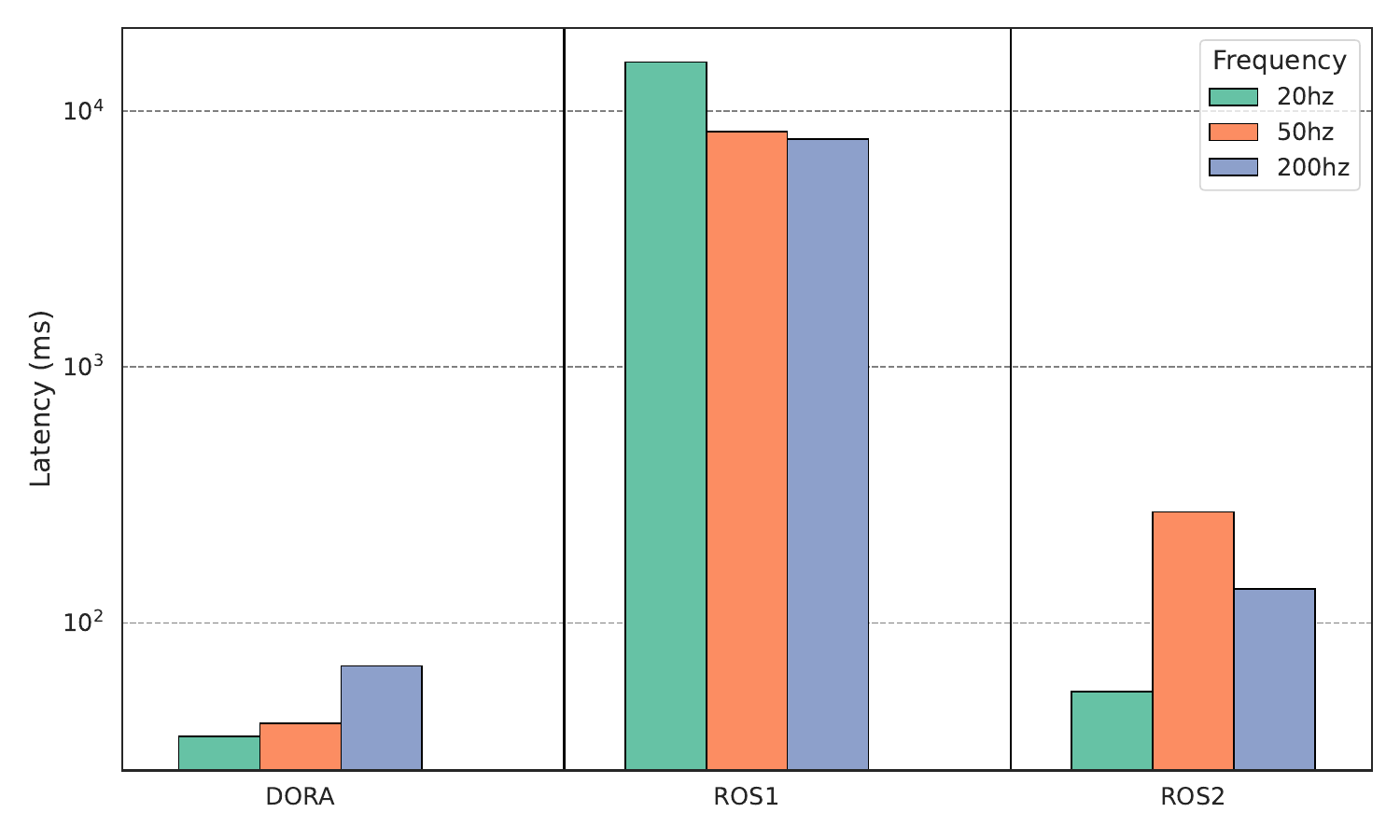}
}
\caption{Impact of transmission frequency on latency}
\label{fig:freq}
\end{figure*}
The results of the second experiment are shown in Figure \ref{fig:freq}. 
Similar to the first experiment, only one producer and one consumer are involved, with the data size fixed at 4 MB, which corresponds to the typical output size of sensors such as RGB cameras and LiDAR. 
The green, orange, and blue lines represent transmission latency at different publication frequencies of 20 Hz, 50 Hz, and 200 Hz, respectively.

In Figure \ref{fig:latency_frequency_local}, it can be observed that, in local transmission, DORA exhibits a slight decrease in latency as the data frequency increases: the latency values are 0.824 ms, 0.784 ms, and 0.728 ms for 20 Hz, 50 Hz, and 200 Hz, respectively. 
This counterintuitive trend --- lower latency at higher frequencies --- is attributed to caching effects and task scheduling optimizations\cite{ros2_latency_overhead}. 
CyberRT shows results consistent with those reported in \cite{oops}, where frequency has little effect on latency. 
However, because its base latency is inherently high, CyberRT’s performance remains significantly worse than that of other robotic middlewares across all frequency settings.
For ROS1, the minimum latency occurs at 50 Hz (12.9 ms), while the maximum latency reaches 146.632 ms at 200 Hz, nearly 11$\times$ higher. 
For ROS2, the minimum latency is observed at 200 Hz (4.947 ms), and the maximum at 20 Hz (17.112 ms), showing a 3$\times$ variation. 

\textit{Overall, in local transmission, DORA consistently demonstrates the highest efficiency, maintaining latencies approximately 21$\times$, 20$\times$, and 6.8$\times$ lower than ROS2 at 20 Hz, 50 Hz, and 200 Hz, respectively.}

In remote transmission (Figure \ref{fig:latency_frequency_remote}), DORA’s latency increases moderately with higher frequencies: 35.935 ms at 20 Hz, 40.517 ms at 50 Hz, and 67.996 ms at 200 Hz. 
For ROS1, latency peaks at 15.575 s at 20 Hz and decreases to 7.792 s at 200 Hz. 
For ROS2, the maximum latency occurs at 50 Hz (271.833 ms), while the minimum is at 20 Hz (54.037 ms). 
\textit{Overall, DORA remains the best-performing system, followed by ROS2, whereas ROS1 performs the worst --- with maximum latencies exceeding 10 seconds, which is completely unacceptable in real-world robotic applications.}

\subsection{Multiple Destinations and Data Fusion}
Robotic applications often involve multi-destination data transmission, where multiple downstream modules require access to the same sensor data. 
For instance, both object detection and image segmentation algorithms may depend on inputs from a single camera node. 
Moreover, multimodal data fusion paradigms also require concurrent access to shared data streams. 
To evaluate DORA’s performance under such conditions, we conducted the multiple destinations and data fusion experiments and compared the results with CyberRT, ROS1 and ROS2. 
Here, multiple destinations represents a scenario with one producer and multiple consumers, while data fusion denotes multiple producers and one consumer. 
During communication, each producer publishes data through a unique topic, and consumers subscribe to this shared topic. 
All experiments were conducted in a local environment.

\begin{figure*}[!t]
\centering
\subfloat[1 publisher and 4 subscribers (1$\to$4)\label{fig:124}]{
    \includegraphics[width=0.48\linewidth]{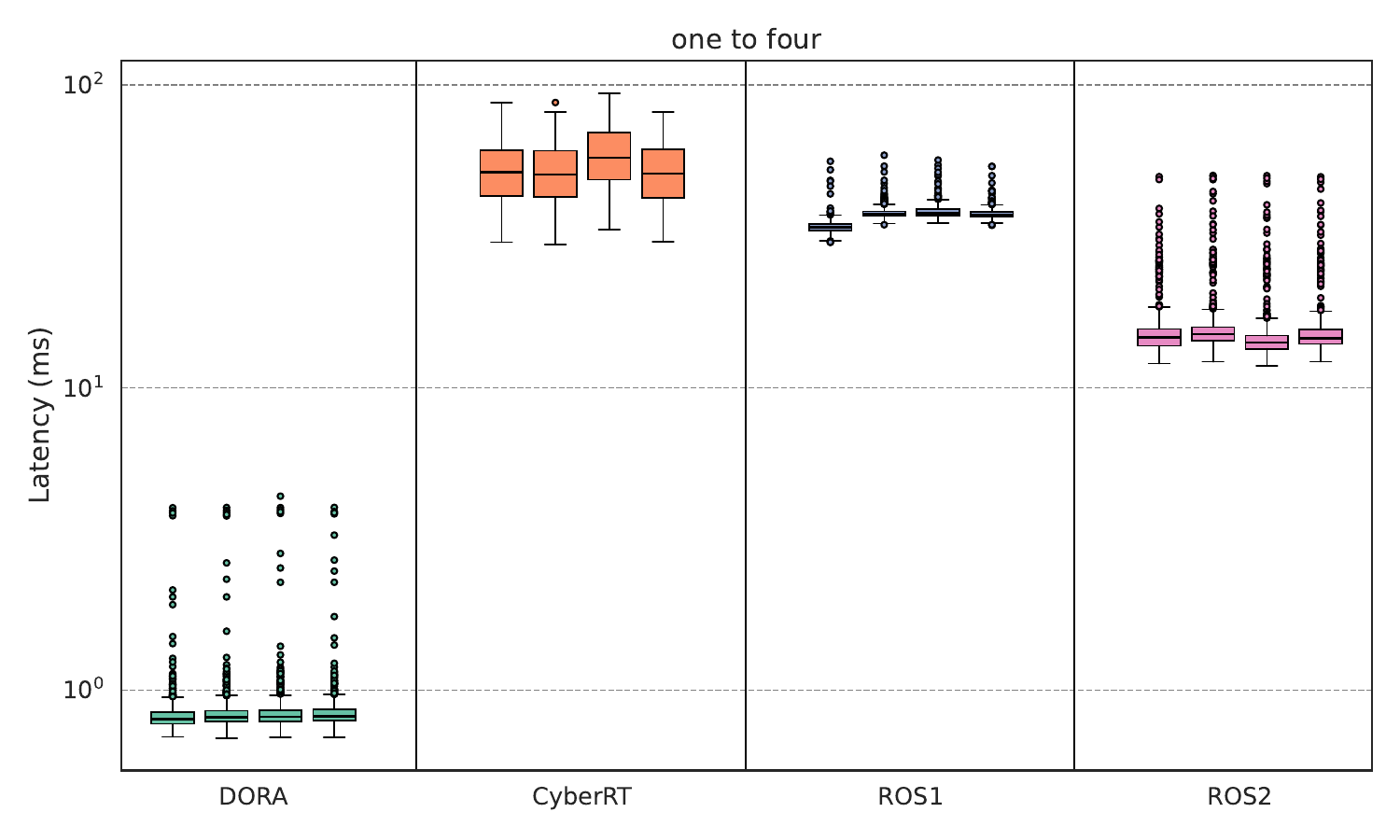}
}
\hfill
\subfloat[1 publisher and 8 subscribers (1$\to$8)\label{fig:128}]{
    \includegraphics[width=0.48\linewidth]{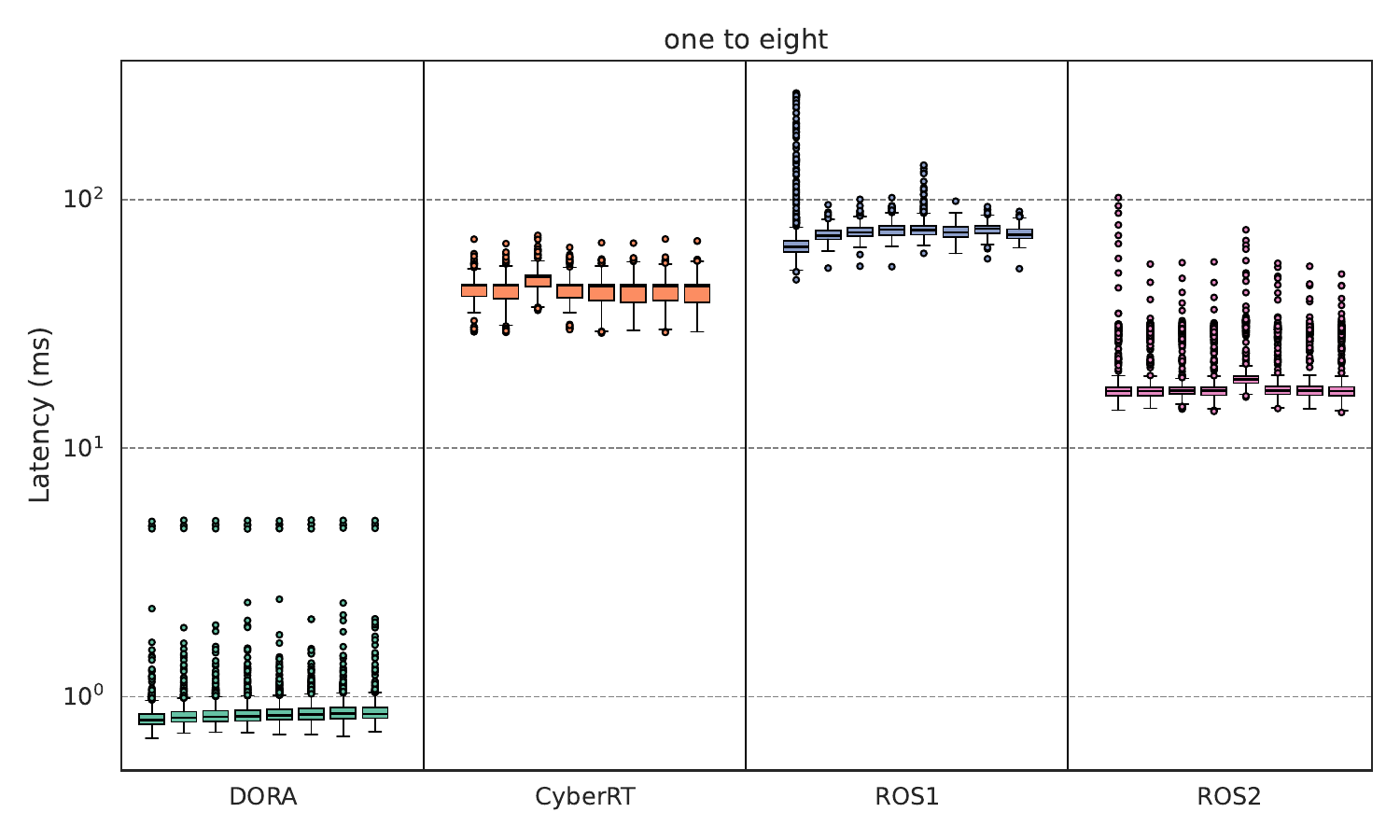}
}

\vspace{0.3cm}
\subfloat[4 publishers and 1 subscriber (4$\to$1)\label{fig:421}]{
    \includegraphics[width=0.48\linewidth]{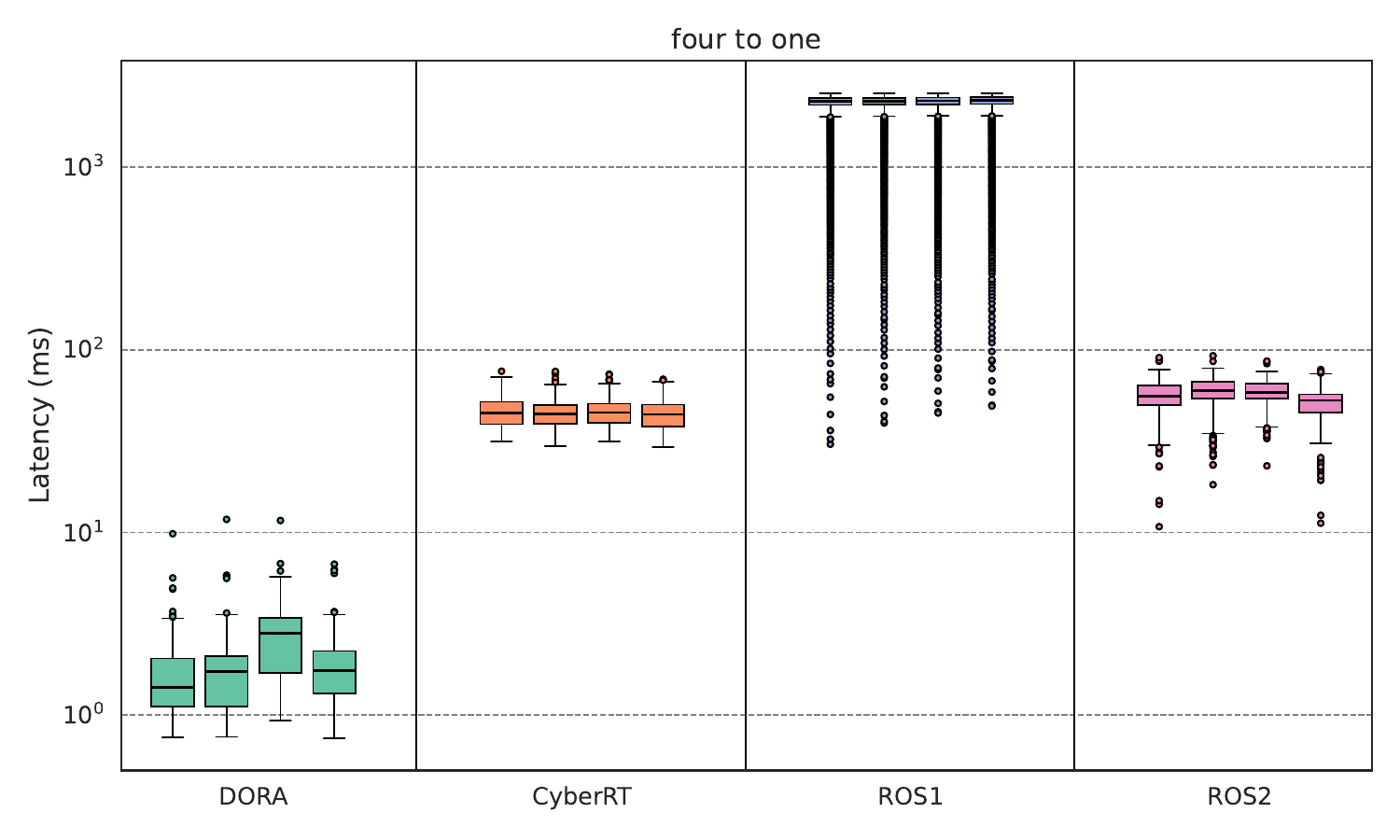}
}
\hfill
\subfloat[8 publishers and 1 subscriber (8$\to$1)\label{fig:821}]{
    \includegraphics[width=0.48\linewidth]{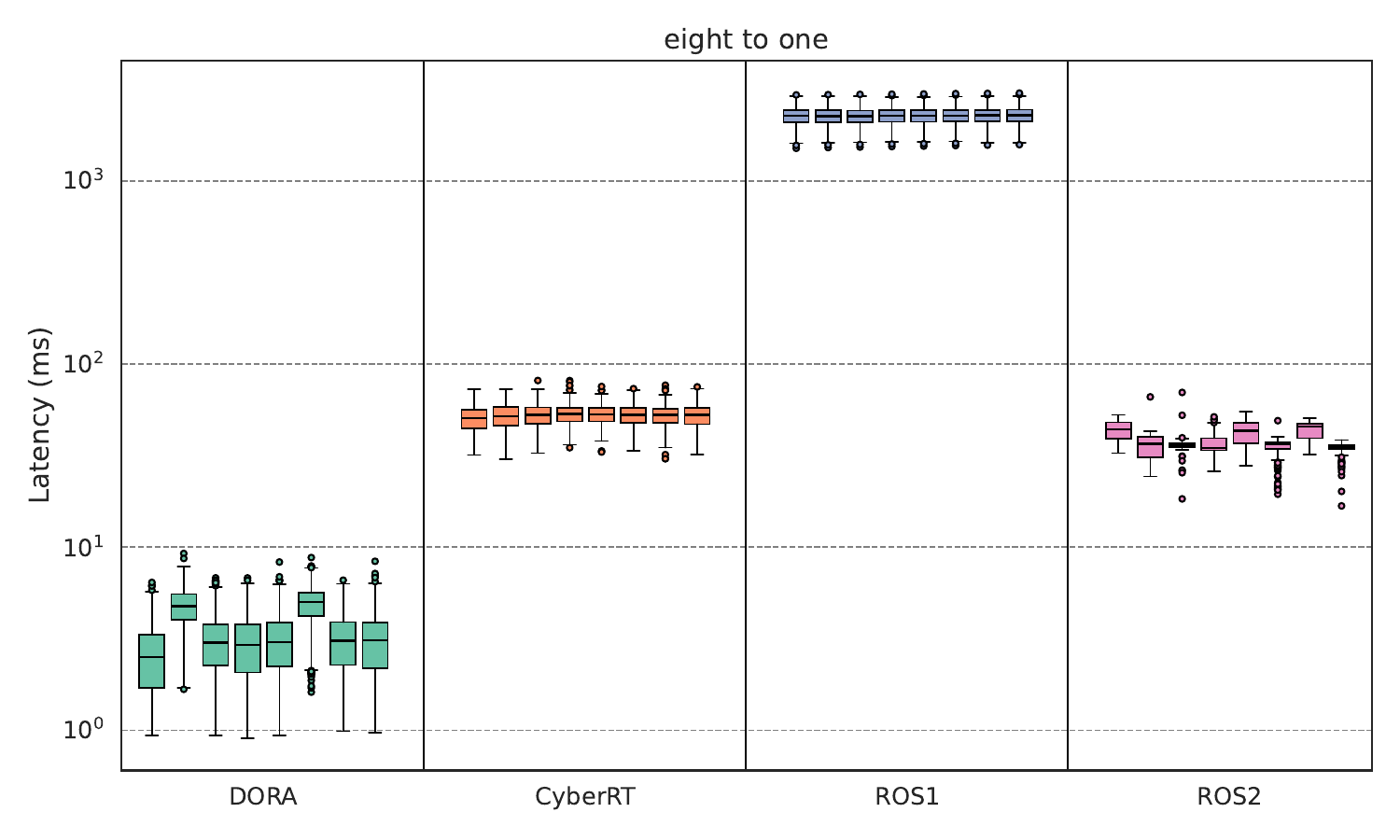}
}
\caption{Mean latency of multiple destinations and data fusion with varying data sizes}
\label{fig:mddf}
\end{figure*}

As shown in Figures \ref{fig:124} and \ref{fig:128}, in the multiple destinations scenario, DORA exhibits significantly lower latency than CyberRT, ROS1 and ROS2. 
The average communication latency for all DORA consumer nodes remains below 1 ms, whereas ROS2 averages around 15.305 ms. 
ROS1 and CyberRT demonstrate notably poor performance --- in the 1$\to$4 setup, their latencies fluctuate around 37.022 ms and 54.05 ms, and in the 1$\to$8 setup, latency surges to nearly 74.52 ms and 43.879 ms, approximately 43.5 and 50 $\times$ higher than that of DORA.
In addition, we evaluated the latency of HPRM\cite{hprm} in the 1$\to$4 setup, which was approximately 10 ms and likewise significantly higher than that of DORA.

Figures \ref{fig:421} and \ref{fig:821} illustrate the performance of DORA, CyberRT, ROS1, and ROS2 in the data fusion scenario. 
It can be observed that although DORA’s latency slightly increases and ranges between 1 ms and 5 ms, it still outperforms all other systems by an order of magnitude. 
In comparison, CyberRT and ROS2 exhibit average latencies of around 50 ms, while ROS1’s latency spikes dramatically to approximately 2.0 s.

Furthermore, across multiple destinations and data fusion scenarios, it can be observed that, similar to other robotic middlewares, DORA maintains consistent latency across all nodes, ensuring communication fairness within the dataflow.

\begin{figure}
    \centering
    \includegraphics[width=\linewidth]{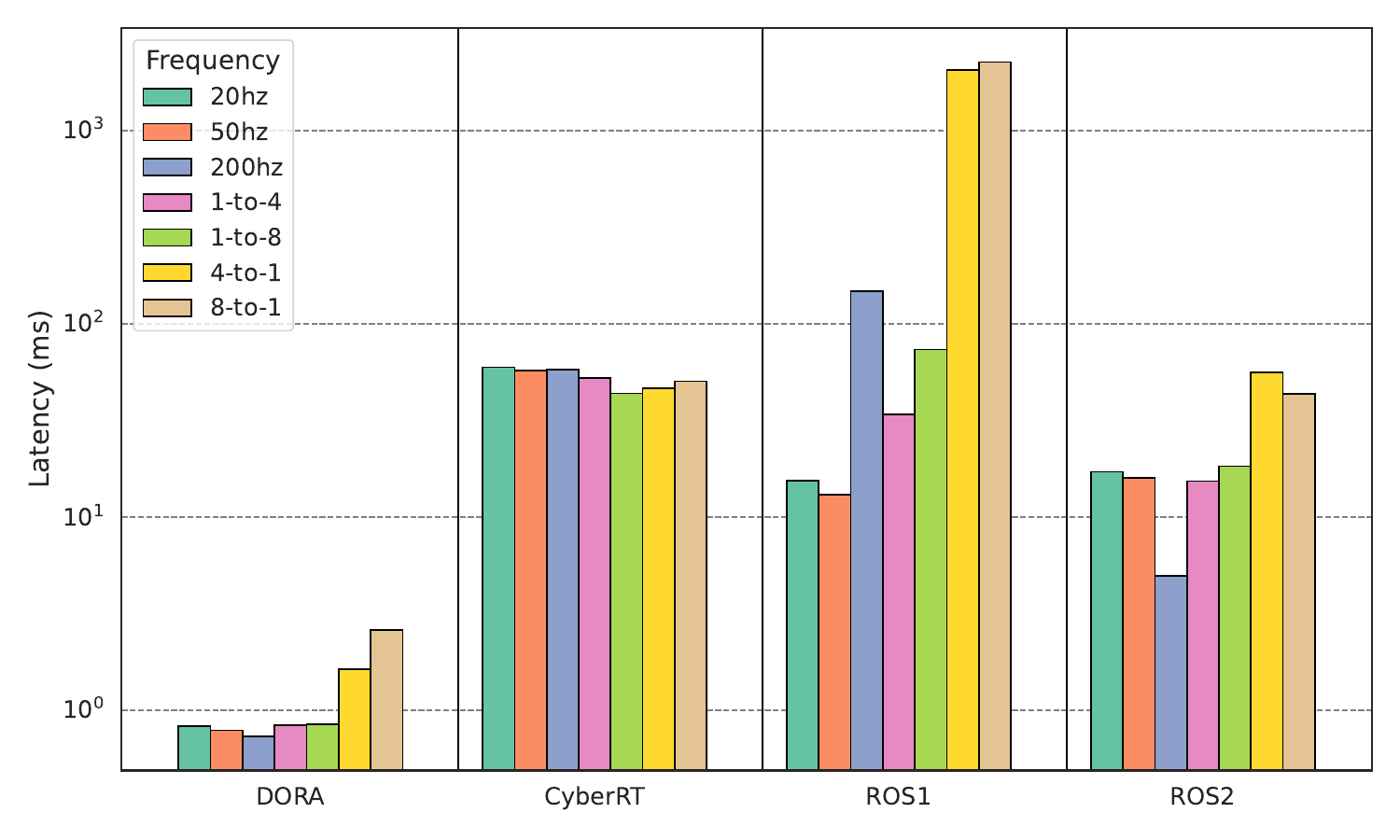}
    \caption{Overall impact of the evaluated scenarios on transmission latency}
    \label{fig:all}
\end{figure}

Finally, we summarized all the above local environment experiments into a single figure, as shown in Figure \ref{fig:all}. 
It can be observed that DORA consistently maintains low and stable transmission latency across various scenarios --- including different data frequencies, multi-destination communication, and data fusion setups.
Although CyberRT and ROS2 also demonstrate stable performance, their latencies remain significantly higher than DORA’s. 
ROS1, on the other hand, shows generally stable latency in most scenarios but experiences a dramatic surge in the N$\to$1 configuration, with latency exceeding about ten times that observed in other cases.

\subsection{Real-World Case Study}

\begin{figure*}[!t]
\centering
\subfloat[DORA with Isaac Sim\label{fig:sim}]{
    \includegraphics[width=0.48\linewidth]{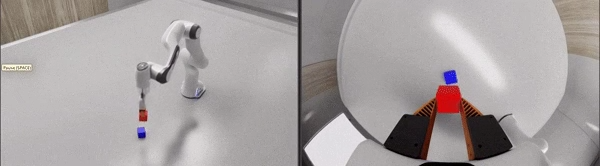}
}
\hfill
\subfloat[Latency Comparison within Isaac Sim\label{fig:DORA_isaacsim}]{
    \includegraphics[width=0.48\linewidth]{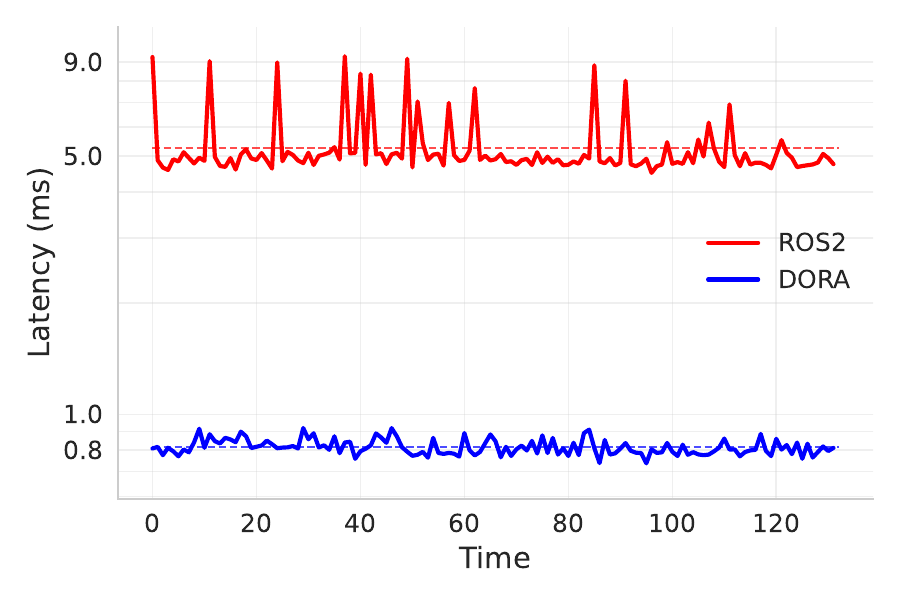}
}
\vspace{0.3cm}
\subfloat[DORA with Realman Gen72 arm\label{fig:real}]{
    \includegraphics[width=0.48\linewidth]{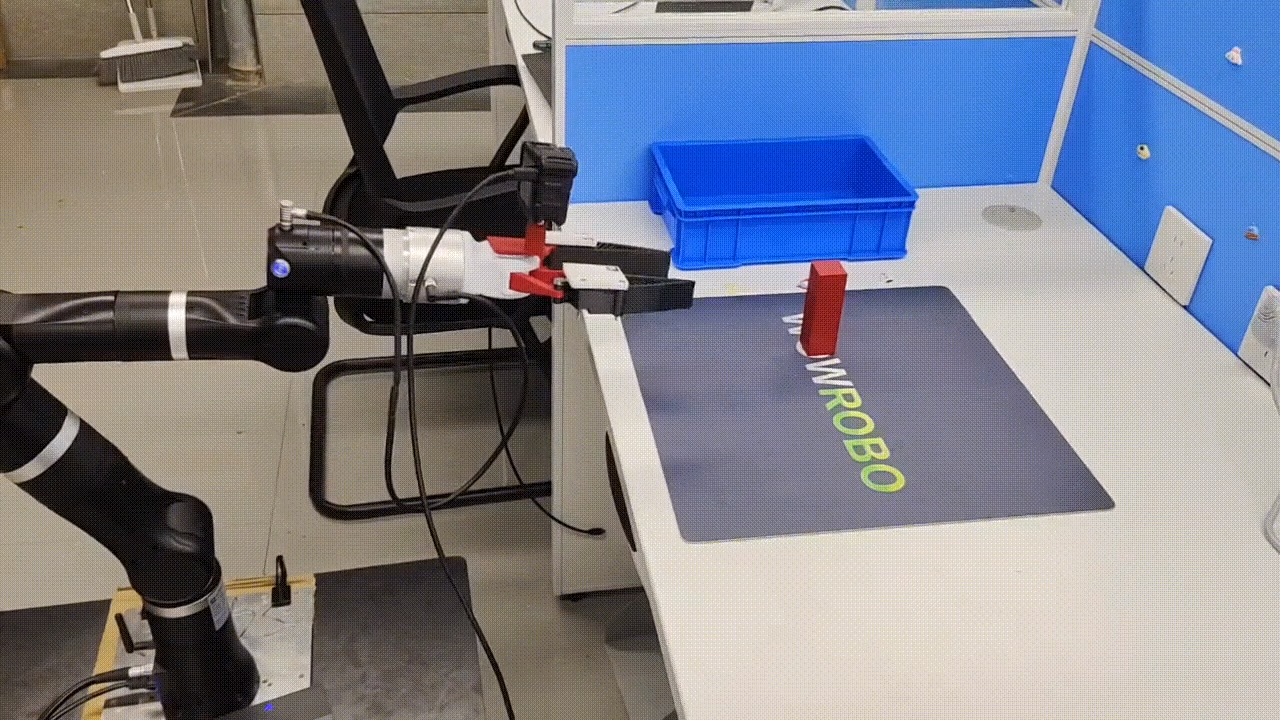}
}
\hfill
\subfloat[Comparison of latency in real-world robot application scenarios\label{fig:DORA_gen72}]{
    \includegraphics[width=0.48\linewidth]{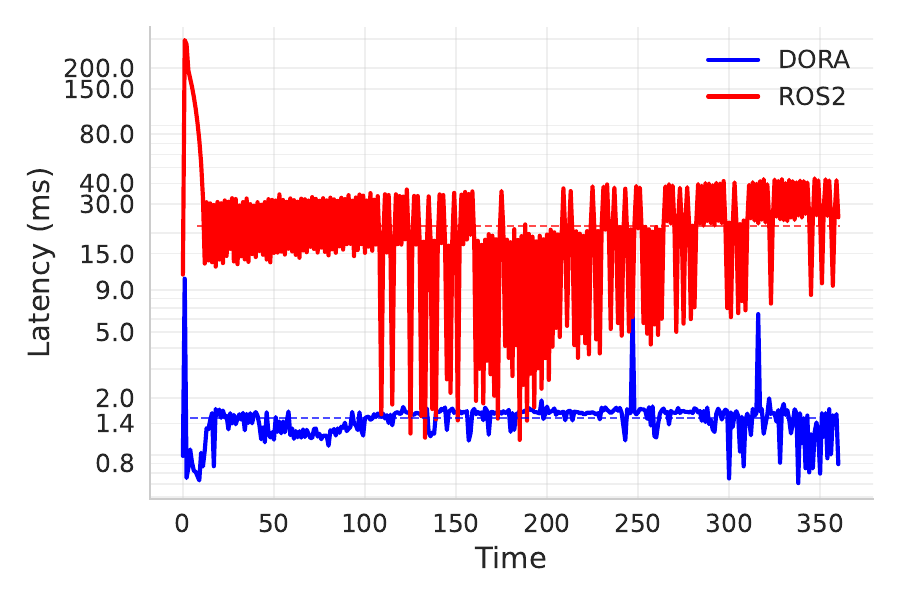}
}
\caption{Real-World case study}
\label{fig:case}
\end{figure*}

To evaluate the practical performance of DORA in robotic applications, we deployed it in two representative settings: a simulation environment and a real robotic arm, measuring both latency and the proportion of each component in the control cycle, as shown in Figure \ref{fig:case}.

In the Isaac Sim\cite{isaacsim} simulation environment, we deployed a Franka robotic arm to perform a stack cube task (Figure \ref{fig:sim}). 
An ACT\cite{aloha} model is used as the inference node, which takes as input the RGB images (640$\times$480, about 0.9 MB) from the wrist camera of the Franka arm and its joint positions obtained from Isaac Sim, and infers the next target joint position to be transmitted back to Isaac Sim for execution. 
Figure \ref{fig:DORA_isaacsim} reports the transmission latency of wrist camera images from Isaac Sim to the inference node during task execution. 
As shown, DORA consistently maintains low latency throughout the process, with an average of approximately 0.8 ms, significantly lower than the average latency of 5.25 ms observed with ROS2.

In the real-world scenario, we deployed a Realman Gen72 robotic arm with ACT model again serving as the inference node (Figure \ref{fig:real}). 
As shown in Figure \ref{fig:DORA_gen72}, DORA still maintains low latency throughout the process, with an average of approximately 1.5 ms, significantly lower than the average latency of 22.0 ms observed with ROS2.

\section{Discussion \& Limitation}
\textbf{Automatic Computation Off-loading}.
With the rapid development of large-scale models such as LLMs, VLMs, and VLAs, the traditional onboard single-machine computing paradigm is increasingly inadequate for the demands of high-performance perception and decision-making in complex environments. 
To address this limitation, it is essential to introduce an edge–cloud collaborative framework into robotic systems—one that ensures real-time responsiveness while fully leveraging the computational power and knowledge resources available in the cloud. 
For complex tasks, this necessitates dynamic task partitioning and migration, enabling the robotic system to exhibit adaptive task deployment capabilities. 
At present, DORA does not yet support automatic computation off-loading, developers must manually define sub-dataflows for the edge and cloud components.

\textbf{Ecosystem Maturity}.
DORA remains in its early developmental stage, with a relatively limited ecosystem of third-party packages and development toolchains. 
Compared with the mature ecosystem of ROS2, DORA still requires broader community participation to expand its capabilities. 
Regarding robot simulation environments, DORA can be easily integrated with MuJoCo due to its decoupled architecture. 
Although the DORA community has developed extensions for NVIDIA Isaac Sim, which primarily provides ROS1/2 interfaces, the performance of these integrations remains relatively limited.

\section{Related Work}
A robotic system typically consists of multiple modular components, each responsible for specific functionalities to ensure system decoupling. 
To coordinate these modules, data is exchanged using either the publish–subscribe or request–response communication pattern. 
However, these conventional communication paradigms are often globally accessible, which introduces potential data leakage risks. 
To address this issue, DORA proposes a dataflow-oriented communication model, in which data transmission is explicitly defined through dataflow graphs, ensuring secure and reliable data exchange.

Early robotic middleware systems relied purely on socket-based communication to handle data transfer within a process, across processes, and between robots. 
Although this approach unified intra- and inter-robot communication, it suffered from high resource consumption and poor latency performance. 
To improve efficiency, ROS2\cite{ros2} and CyberRT\cite{CyberRT} adopted DDS as their communication backbone. 
ROS2 relies entirely on DDS, while CyberRT employs shared memory for IPC and uses DDS for remote communication. 
However, both architectures still involve redundant data copying, which degrades overall performance. 
Moreover, their complex (de)serialization protocols, though enabling interoperability among heterogeneous robotic systems, come at the cost of significant computational and memory overhead.
CyberRT 10.0 introduced the Arena mechanism, which allocates transmitted data directly in predefined shared memory regions, thereby reducing intermediate data movement. 
Nonetheless, this approach requires large contiguous shared memory allocations, limiting its flexibility.
RobustZ\cite{robustz} and Zoro\cite{zoro}, on the other hand, focus on the reliability challenges posed by shared memory. 
They propose separating control data from communication data and transmitting them through distinct channels. 
This design partially improves communication latency and system reliability, yet it does not fundamentally eliminate the overhead caused by (de)serialization.
In this work, DORA introduces a unified data representation format that eliminates deserialization overhead, along with an optimized shared-memory allocation and reclamation algorithm tailored to robotic workloads, enabling highly efficient and low-latency data transmission.

\section{Conclusion}
In this study, we present DORA, a dataflow-oriented and high-performance robotic middleware. 
From development paradigm perspective, DORA adopts an intuitive and transparent dataflow graph as the blueprint for the entire robotic application. 
Built upon this model, we designed \texttt{Dora-Coordinator} and \texttt{Dora-Daemon} to manage global and local dataflows, respectively, enabling efficient and secure intra-robot communication while also supporting multi-robots collaboration and edge-cloud cooperation.
To address the communication performance limitations of existing robotic middleware, DORA introduces an in-memory object store that eliminates deserialization in local message transmission, achieving zero-copy communication and reduced resource consumption. 
Moreover, an on-demand shared-memory allocation algorithm avoids message fragmentation and reassembly, further reducing communication latency.
Comprehensive experiments and analyses demonstrate that: 1) Compared with ROS2 and CyberRT, which exhibit over 30\% CPU utilization, DORA’s deserialization overhead is almost 0; 2) For both large-scale intra-robot and remote communication, DORA maintains transmission latency within 3 ms and 90 ms, respectively; 3) Under complex communication scenarios, DORA consistently achieves superior fairness and performance. 
It is worth noting that DORA is progressively building an open robotics community, dedicated to providing a high-performance and developer-friendly middleware platform for the broader robotics ecosystem.

\bibliographystyle{IEEEtran}
\bibliography{reference}

\vfill

\end{document}